%% file: eacl2023.tex
\pdfoutput=1

\documentclass[11pt]{article}

\usepackage[]{naacl2021}

\usepackage{times}
\usepackage{latexsym}

\usepackage[T1]{fontenc}

\usepackage{microtype}

\usepackage{inconsolata}
\usepackage{amsmath}
\usepackage{graphicx}
\usepackage{multirow}
\usepackage{soul}
\usepackage{pifont}
\usepackage{float}
\usepackage{times}
\usepackage{latexsym}
\usepackage{graphicx}
\usepackage{booktabs}
\usepackage{subcaption}
\usepackage[ruled,vlined]{algorithm2e}
\usepackage{multirow}
\usepackage{tabularx}
\usepackage{amsmath}
\usepackage{bbold}
\usepackage{mathtools}
\usepackage{xcolor}
\usepackage{booktabs}
\usepackage{tabularx,ragged2e}
\usepackage{enumitem}
\usepackage{xcolor}
\definecolor{darkgreen}{rgb}{0,0.5,0}

\newcommand{\name}{\textsc{I}nstruct\textsc{ABSA}\xspace}

\title{ \name{}: Instruction Learning for Aspect Based Sentiment Analysis}

\author{Kevin Scaria \hspace{9pt} Himanshu Gupta \hspace{9pt} Siddharth Goyal\hspace{9pt} \\ \textbf{Saurabh Arjun Sawant \hspace{9pt} Swaroop Mishra \hspace{9pt}Chitta Baral}\\
  Arizona State University \\
\tt\small   \texttt{\{kscaria, hgupta35\}}@asu.edu
  }

\begin{document}
\maketitle
\begin{abstract}
We introduce \name{}, an instruction learning paradigm for Aspect-Based Sentiment Analysis (ABSA) subtasks.
Our method introduces positive, negative, and neutral examples to each training sample, and instruction tune the model (T$k$-Instruct) for ABSA subtasks, yielding significant performance improvements.
Experimental results on the Sem Eval 2014, 15, and 16 datasets demonstrate that \name{} outperforms the previous state-of-the-art (SOTA) approaches on Term Extraction (ATE), Sentiment Classification(ATSC) and Sentiment Pair Extraction (ASPE) subtasks.
In particular, \name{} outperforms the previous state-of-the-art (SOTA) on the Rest14 ATE subtask by 5.69\% points, the Rest15 ATSC subtask by 9.59\% points, and the Lapt14 AOPE subtask by 3.37\% points, surpassing 7x larger models.
We also get competitive results on AOOE, AOPE, and AOSTE subtasks indicating strong generalization ability to all subtasks. 
Exploring sample efficiency reveals that just 50\% train data is required to get competitive results with other instruction tuning approaches. 
Lastly, we assess the quality of instructions and observe that \name{}'s performance experiences a decline of $\sim10\%$ when adding misleading examples
\footnote{Experiments and results are available at 
\url{https://anonymous.4open.science/r/InstructABSA-EB71}
}.
\end{abstract}

\section{Introduction}
\begin{figure}[t!]
	\centering
	\includegraphics[width= \linewidth, height= 5.9 cm]{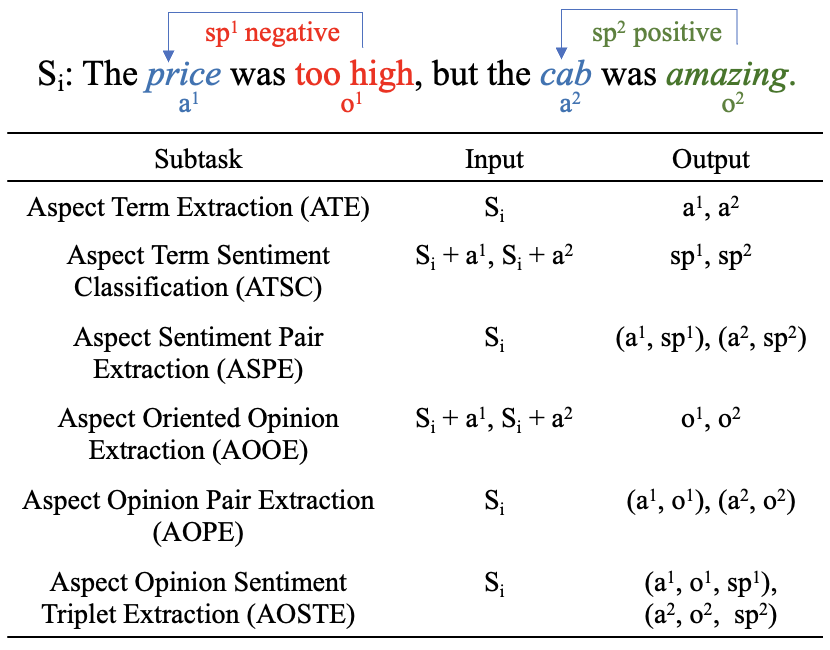}
	\caption{Illustration of the six ABSA subtasks where $S_i$ is the $i^{th}$ sentence, $a^i$ are the aspect terms, $sp^i$ are the sentiment polarities and $o^i$ is the opinion terms.}
	\label{fig:teaser}
\end{figure}

Aspect Based Sentiment Analysis (ABSA) plays a vital role in understanding the fine-grained sentiments expressed by users \cite{Zhang2012SentimentAA}. As illustrated in Figure \ref{fig:teaser}, ABSA extracts aspects and classifies the aspect's sentiment polarity by extracting and understanding the author's opinions. 
Instruction learning paradigm \cite{mishra-etal-2022-cross, wei2022finetuned,gupta2023instruction} has significantly improved the reasoning abilities of large language models (LLMs) and has shown impressive results across various tasks \cite{wang2022self,lu2022learn}.  
Owing to its previous success, we propose \name{}, instruction learning for aspect based sentiment analysis (ABSA).  
Our approach involves further instruction tuning of the T$k$-Instruct model \cite{wang-etal-2022-super} to address six subtasks of ABSA as shown in figure \ref{fig:teaser}. We add instruction prompts specific to the downstream ABSA subtasks in the form of task definitions, followed by positive, negative, and neutral examples.

 We carried out extensive experiments on the SemEval 2014, 15, and 16 datasets \cite{pontiki-etal-2014-semeval,Pontiki2015SemEval2015T1,Pontiki2016SemEval2016T5}, and the dataset by \cite{Peng_Xu_Bing_Huang_Lu_Si_2020} for the AOSTE subask, which comprises the laptops and restaurants domain. 
 Across the subtasks in both domains, \name{} outperforms SOTA approaches. 
Specifically, for the 2014 ATE subtask, we obtain F1-score of 92.3 and 92.76 (Lapt14, Rest14), surpassing SOTA by $4.37\%$ and $5.69\%$ points respectively.
For the ATSC subtask, \name{} attains an accuracy of 84.50 in the Rest15 dataset exceeding the previous results by $9.59\%$ points. 
In the Rest14 dataset ATSC subtask, our approach gets a competitive accuracy score of 86.25 compared to the SOTA of 90.86.
For the ASPE subtask, \name{} achieves F1-score of 79.34 and 79.47 (Lapt14, Rest14), outperforming SOTA by $3.37\%$ and $1.4\%$ points, respectively.
We get competitive results on AOOE and AOSTE approaches as well (\S \ref{sec_results}). 

We conduct a thorough analysis along several lines of enquiry. 
We showcase sample efficiency of \name{} by achieving competitive scores using roughly 20\% of training samples as compared to \citet{varia2023instruction}'s instruction tuning approach.
We compare \name{} with finetuning methods such as Low-Rank Adaptation (LoRA) \cite{hu2021lora} to find that there is a sizebale gap of $\sim20\%$.
To understand the effect of different instructions for ABSA, we change the prompts on the lines of definition and task manipulation. 
We find that delusive examples roughly decrease the approaches results by $\sim10\%$ giving a strong evidence of the impact of instructions on \name{}.
We also provide evidence of cross-domain and joint-domain generalizations arising as part of our proposed approach. 

\noindent\textbf{Contributions:}(a) we introduce \name{}, which achieves performance gains on ABSA subtasks of SemEval 2014,15 and 16 datasets, surpassing the previous SOTA models. 
(b) Despite using a 200M model, \name{} outperforms or get competitive results over the prior SOTA models with 1.5B parameters. 
(c) Finally, we provide an analysis of the impact of our method in terms of sample efficiency, adapter methods, effect of instruction and domain generalization.

\section{\name{}: Instruction Learning for ABSA }

We describe the mathematical formulation of ABSA subtasks and the proposed approach.

Let $S_i$ represent the $i^{th}$ review sentence in the training sample, where $S_i = {w_{i}^1, w_{i}^2, ..., w_{i}^n}$ with $n$ as the number of tokens in the sentence. 
Each $S_i$ contains a set of aspect terms denoted by $A_i = {a_{i}^1, a_{i}^2, ..., a_{i}^m} | m \le n$, and the corresponding opinion terms and sentiment polarities for each aspect term are denoted by $O_{i} = {o_{i}^1, o_{i}^2, ..., o_{i}^m}$ and $SP_{i} = {sp_{i}^1, sp_{i}^2, ..., sp_{i}^m}$ respectively, where $sp_i^k \in [ positive, negative, neutral ]$. 
The ABSA tasks are described as follows:\\
ATE: $A_i = LM_{ATE}(S_i)$\\
ATSC: $sp_i^k = LM_{ATSC}(S_i, a_i^k)$\\
ASPE: $[A_i, SP_i] = LM_{ASPE}(S_i)$\\
AOOE: $o_{i}^k = LM_{AOOE}(S_i, a_i^k)$\\
AOPE: $[A_i, O_i] = LM_{AOPE}(S_i)$\\
AOSTE: $[A_i, O_i, SP_i] = LM_{AOSTE}(S_i)$\\
In these equations, $LM$ represents the language model, and the corresponding inputs and outputs are defined accordingly. As part of our approach, we instruction tune $LM_{subtask}$ by prepending task-specific prompts to each input sample to arrive at $LM_{subtask}^{Inst}$ (Details in \S \ref{sec:prompt_examples}). 



\section{Results and Analysis}
\label{sec_results}

\subsection{Sub Task Results}

\input{tables/ATERes.tex}
\input{tables/ATSCRes.tex}

Tables \ref{tab:ate_result} - \ref{tab:aoste_result} denotes the results of ATE, ATSC, ASPE, AOOE, AOPE and AOSTE subtasks respectively. 
All the results reported are the average values from 5 runs for each experiment.
For \textbf{ATE} subtask (Table \ref{tab:ate_result}), \name{} surpasses SOTA on Lapt14, Rest14, 15, and 16 datasets surpassing 7x larger models (\citet{hosseini-asl-etal-2022-generative} uses GPT-2 with 1.5B parameters).
For \textbf{ATSC} subtask, \name{}-2 achieves SOTA of Rest 15 while remaining competitive of Lapt and Rest 14 dataset.
For the \textbf{ASPE subtask} (Table \ref{tab:joint_result}), \name{} acheives SOTA for all four datasets. 
In the \textbf{AOOE} subtask (Table \ref{tab:aooe_result}) \name{} achieves an F1 score of 76.42 and 77.16 for the Lapt14 dataset, outperforming IOG and ONG. 

In the \textbf{AOPE task} subtask (Table \ref{tab:aope_result}), \name{} suffers compared to the existing models. 
For the \textbf{AOSTE task} (Table \ref{tab:aoste_result}), Seq2Path achieves the highest F1 scores for the datasets, however, our models achieve competitive results for Rest14. 
The performance of \name{} in AOPE and AOSTE is subpar due to exposure bias. 
For sentiment pair extraction tasks, the model had to decode only the aspect terms followed by sentiments that were constrained to positive, negative, and neutral labels. 
However, for the opinion pair extraction tasks and triplet extraction tasks, the model suffers higher exposure bias since the opinion terms are not grounded and could potentially be any word in the vocabulary \cite{zhang-etal-2020-minimize}.

\input{tables/JointTaskRes.tex}
\input{tables/AOOE_res}
\input{tables/AOPE_res}
\input{tables/AOSTE_res}

\subsection{Analysis}
\label{analysis}
In this subsection, we analyze InstructABSA on multiple line of enquiries.


\paragraph{Cross-Domain and Joint Domain Evaluation:} 
 
In cross domain setting, we train the model on a train set from one domain and test on test set from another domain. 
In joint domain setting, the train data of the domains (laptops and restaurants) are combined to train the model, and it is evaluated on both test sets. 
Both experiments are performed on ATE, ATSC and ASPE subtasks for both instruction-tuned models (\name{}-1 \& 2).
Table \ref{tab:cross_domain} presents the cross domain experiment results. 
When trained on Lapt14 and tested on Rest14, \name{}-1 shows a drop in F1-score for the ATE and Joint Task compared to \name{}-2.
For the ATSC task, similar trends were obtained with an accuracy of 75.53 from \name{}-1 and 80.56 from \name{}-2.
The joint domain experiments are present in Table \ref{tab:cross_domain_joint}.
The availability of additional training data for ATE subtask helps the language models as the proposed model surpasses the previously achieved SOTA.

\input{tables/crossdomainRes.tex}

\input{tables/cross_domain_joint.tex}

\input{tables/LoraComparison}

\paragraph{Delusive examples reduce \name{}'s performance} 

\begin{figure}[t!]
	\centering
	\includegraphics[width= \linewidth, height= 6.1 cm]{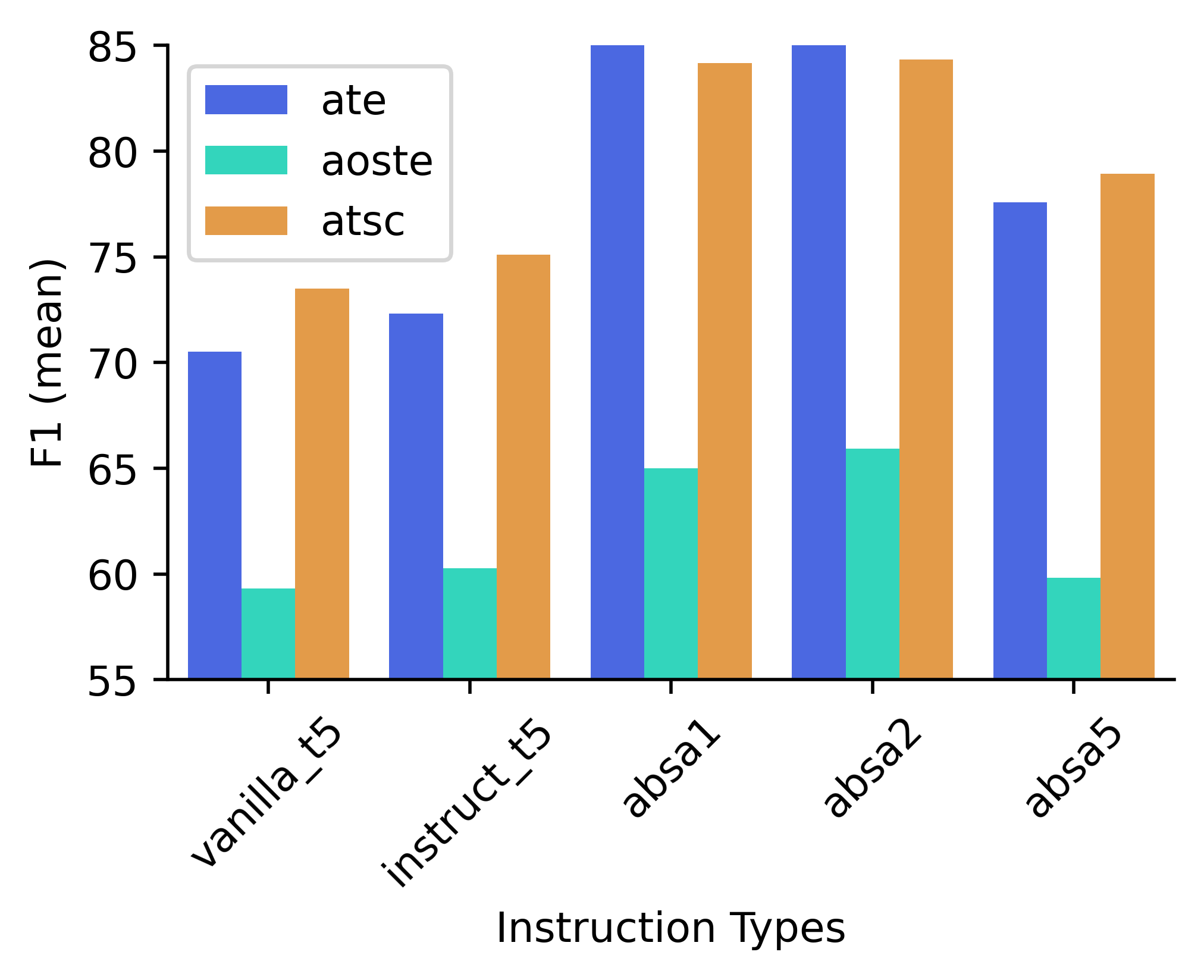}
	\caption{Comparison of various instruction configuration and its performance on ATE, AOSTE and ATSC subtasks. 
 vanilla\_t5 and instruct\_t5 represent the base T5 model with and without instruction tuning on the dataset. 
 absa1 includes a definition followed by 2 positive exemplars, absa2 includes a definition followed by 2 positive, negative, and neutral examples, and finally, absa5 is the delusive configuration with incorrect input and output mappings respectively.}
	\label{fig:eff_instr}
\end{figure}

We analyze the impact of instruction tuning along the lines of experiments proposed by \citet{kung2023models}, focusing on task definition and example manipulation. 
In task definition manipulation, we explore original, simplified, and empty definitions, but only use the empty configuration with vanilla T5 and T$k$-instruct models. 
In task example manipulation, we study original, delusive, and empty examples, as well as additional configurations. 
Detailed results can be found in Figure \ref{fig:eff_instr} and Tables \ref{tab:instr_eff_ate}, \ref{tab:instr_eff_atsc}, and \ref{tab:instr_eff_aoste}. 
Notably, \name{}-1 and 2 outperform the vanilla models, highlighting the effectiveness of instruction tuning for most ABSA subtasks.

\begin{figure}[t!]
	\centering
	\includegraphics[width= \linewidth, height= 6.25 cm]{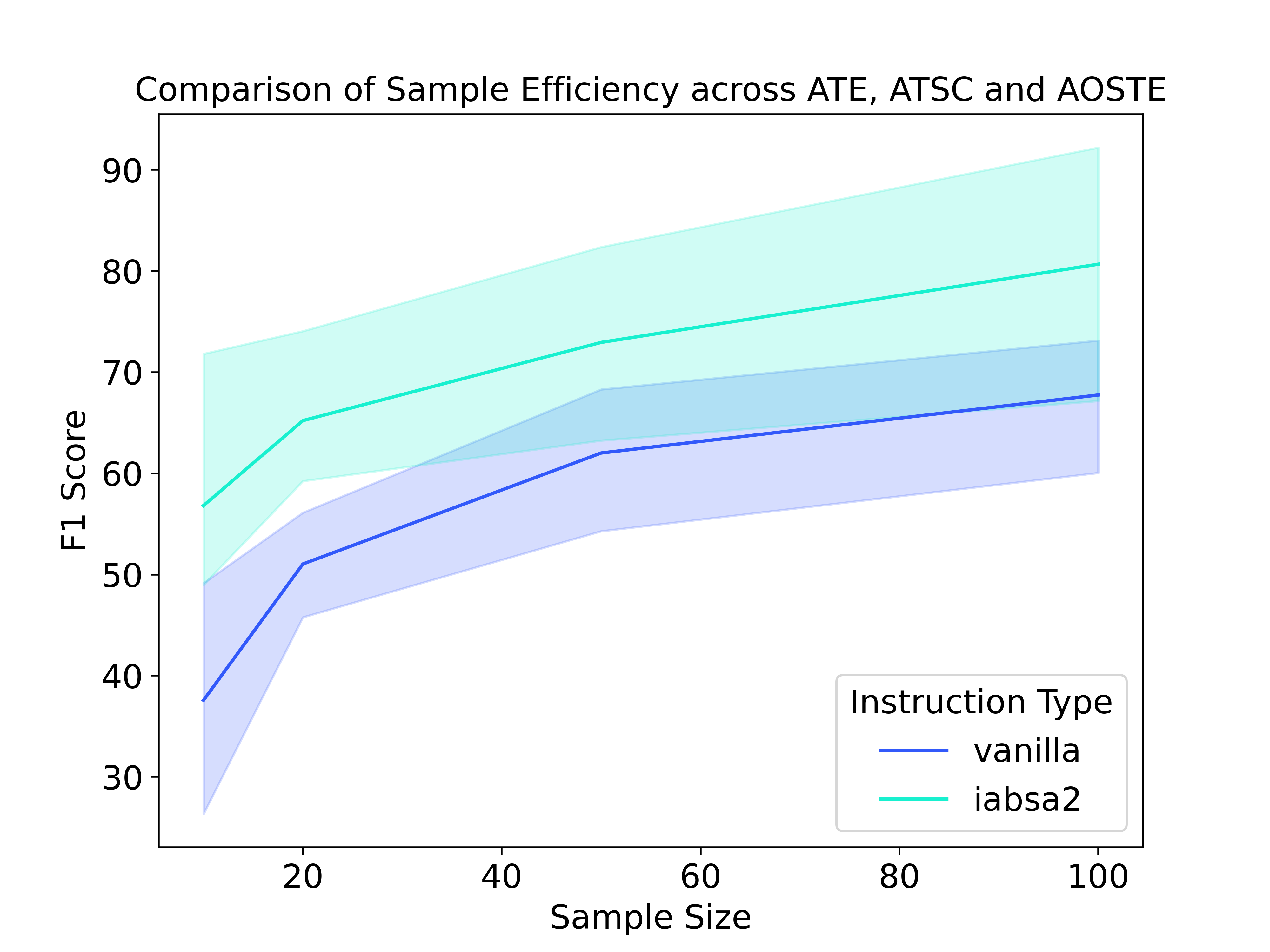}
	\caption{Comparison of sample efficiency on ATE, AOSTE and ATSC subtasks between \name{}-2 and vanilla model. Sample size is \% of training data.}
	\label{fig:sample_eff}
\end{figure}

\paragraph{Competitive scores with just 50\% train samples}
\citet{gupta2023instruction} showcased the effects of sample efficiency via instruction tuning. 
Following that work, we explore the performance of instruction tuning by using a smaller percentage of the training set. 
We carry out experiments to identify the sample efficiency gains for ABSA subtasks. 
The results are presented in Figure \ref{fig:sample_eff} and Table \ref{tab:sample_efficiency}. 
We get competitive scores with our best scores when using roughly 50\% train samples, demonstrating sample efficiency of \name{}. 

Figure \ref{fig:sample_eff} also showcases the performance of the vanilla T5 base model finetuned with the same number of samples.
As shown in the figure, the vanilla model's performance is consistently lower compared to \name{}.

\paragraph{Adapter methods leading to poor performance} 
 We compare the performance of parameter efficient finetuning method Low-Rank Adaptation (LoRA)\cite{hu2021lora} with our instruction tuning approach \name{}. LoRA can lead to significant improvements in memory efficiency and computational efficiency, but it can also lead to a drop in performance.
The experiment is performed on all the subtasks, and the results are presented in Table \ref{tab:lora_comparison}.
As seen in the table a drop of 13.32\% points in ATE, 26.8\% points in ATSC and 19.8\% points in ASPE. 
The drop in scores is significant to overlook when aiming to reap the advantages of a computationally optimized finetuning method.

\section{Conclusion}
We proposed \name{}, an instruction-tuned modeling approach for all subtasks of ABSA. 
Our findings show that \name{} surpassed the previous scores on several tasks and achieved competitive scores on the rest using a significantly smaller model than previous approaches. 
We further analyzed the performance of the approach along several lines of enquiry revealing several interesting findings. 
Finally, we release our code and hope that our work will encourage further research in this direction.

\section*{Limitations}
Our study is limited to the Sem Eval 2014, 15, and 16 datasets, that are widely used in recent works. 
Future studies should include the extension of this work on other ABSA datasets to test the generalizability of our findings.
We conducted our experiments using a 200M model, which may limit the applicability of our findings to smaller models. 
Future studies could consider using even smaller instruction-tuned models to analyze their performance.
Our study was conducted using T$k$-Instruct models for the English language. 
As a result, our findings may not be directly applicable to other languages. 
Future studies should include a multilingual dataset and a multilingual instruction-tuned model to investigate the model's performance across different languages.
Future studies could consider using even smaller instruction-tuned models to analyze their performance.
Our study was conducted using T$k$-Instruct models for the English language. 
As a result, our findings may not be directly applicable to other languages. 

\section*{Ethical Considerations}
We acknowledge that the T5 model used in our experiments may have inherent biases due to the pre-training and instruction-tuning data used. 
While stress testing was not conducted, we believe that from our research no additional issues arise related to privacy, fairness, bias, and discrimination.
We 
Our work directly contributes to the topic of aspect based sentiment analysis and we believe that our work will have a positive impact on the scientific community. 
We remain dedicated to advancing the responsible use of AI and will continue to prioritize ethical considerations in all our future research endeavors.

\bibliography{anthology,custom}
\bibliographystyle{acl_natbib}

\clearpage

\section*{Appendix}
\appendix

\section{Choosing Samples as Instruction Exemplars:} 
From Table \ref{tab:dataset_description_1}, it can be noticed that the distribution of count of aspects across Lapt14, Rest14, Rest15, and Rest16 datasets is centered around one, two, and three aspects which account for 30\%, 11\%, and 4.5\% of total aspects. 
Thus for our instruction exemplars, we randomly select samples that have aspects ranging between 1 and 3. 
We exclude these exemplars during evaluation.

\section{Instruction Effectiveness Study}
To validate the effect of instruction tuning on the performance of various ABSA sub tasks, 
We analyse effect of instruction tuning along the lines of experiments proposed by \citet{kung2023models}. 
We carry out our analysis on two aspects: task definition manipulation and task example manipulation. 
In \emph{task definition manipulation}, controlled experiments are conducted to examine whether models truly comprehend and utilize the semantic meaning of task definitions. 
Three levels of granularity was proposed viz. \emph{original}, \emph{simplified}, and \emph{empty}. 
The simplified version removes all semantic components from the task definition, leaving only the output space information. 
The empty version eliminates the task definition altogether. 
However, as part of the task definition manipulation experiment we only conduct the empty configuration with vanilla\_t5 and vanilla\_tk where t5 is the T5-base model and tk is the T$k$-instruct base model. 
In \emph{task example manipulation}, the influence of task examples on model learning is investigated. 
Three types of task examples are compared: \emph{original}, \emph{delusive}, and \emph{empty}. 
The original setup includes one/two positive example (absa1), while the delusive examples consist of negative examples with incorrect input-output mappings (absa6). 
The empty setup excludes task examples during training (task\_def\_only). 
We additionally carry out different configuration of task examples and call it additions, where we add 2 positive, negative and neutral examples (absa2), 2 negative (absa3), 2 neutral (absa4) and 1 positive, negative and neutral example (absa5). 
The detailed reports are presented in the Figure \ref{fig:eff_instr} and Tables \ref{tab:instr_eff_ate}, \ref{tab:instr_eff_atsc} and \ref{tab:instr_eff_aoste} . 
It is evident that for most ABSA subtasks, the instruction configuration of \name{}-1 and 2 yields the best performance. 
Additionally, it can be seen that both the vanilla models do not give the best results solidifying the effectiveness of further instruction tuning. 

\begin{figure*}[ht!]
	\centering
	\includegraphics[width= \linewidth, height= 6.1 cm]{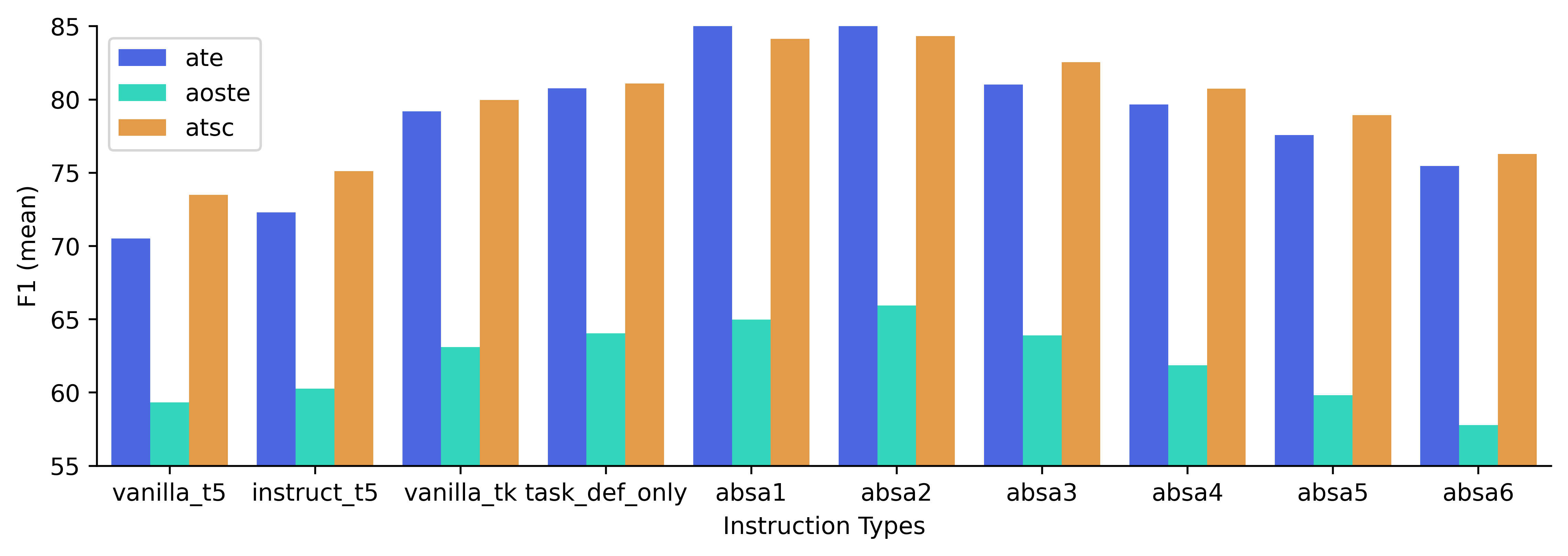}
	\caption{Comparison of various instruction configuration and its performance on ATE, AOSTE and ATSC subtasks. 
 Vanilla\_t5 and Vanilla\_tk represent the models trained without any instruction. 
 absa1, absa2, absa3, absa4, absa5 are different instruction configurations that include a definition followed by 2 positive, 2 positive, negative and neutral examples, 2 negative examples, 2 neutral examples, 1 positive, negative and neutral examples and finally examples with incorrect input and output mappings respectively. task\_def\_only only contains the task definitions.}
	\label{fig:eff_instr}
\end{figure*} 

\section{Detailed Dataset Description:}
\label{sec:dataset}

\input{tables/ATSCData.tex}
\input{tables/JTData.tex}

Table \ref{tab:dataset_description_1} displays the dataset description with respect to the count of aspect terms for all subtasks. 
For the training set, 1557 reviews in Lapt14 and 1020 reviews in Rest14 have no aspect terms and their corresponding polarities. 
Similarly, in the test set, 378 reviews in Lapt14 and 194 reviews in the Rest14 have no aspect terms and corresponding polarities. 
The dataset description for the ATSC subtask is presented in Table \ref{tab:dataset_description_2}. 
To maintain consistency with the previous approaches for the ATSC task, we also ignore conflict labels.

\section{Experimental Setup}
We use the T$k$-Instruct-base-def-pos as the instruction-tuned model $LM_{Inst}$.
We use two configurations of instructions as prompts for our experiments. 
\name{}-1 has the instruction prompt that includes the definition of the ABSA subtasks followed by 2 positive examples for the respective task.
\name{}-2 has the definition followed by 2 positive, negative, and neutral examples.

\paragraph{Dataset:} SemEval 2014,15 and 16 datasets are used for our experimentation. 
The dataset is used as a benchmark for ABSA tasks and has customer reviews from three domains; laptops (Lapt14), hotels (Hotel15), and restaurants (Rest14, Rest15, and Rest16).
More details can be found in \S \ref{sec:dataset}. 

\paragraph{Hyperparameters} 
Model: T$k$-Instruct-base-def-pos \footnote{\url{https://huggingface.co/allenai/tk-instruct-base-def-pos}},
GPU: 1xNvidia Tesla P40,
Train Batch Size: 16 for ATE and ATSC, 8 for other subtasks.
Gradient Accumulation Steps: 2,
Initial learning rate: 5e-5,
Num of Epochs: 4

\paragraph{Evaluation Metric:} Following previous approaches \cite{zhang-etal-2021-towards-generative, luo-etal-2020-grace}, we use the F1-score for ATE, AOPE, AOOE, AOPE, AOSTE, and the accuracy metric for ATSC subtask.

\section{Extended Related Work}

LMs and deep learning methods have been used for a plethora of downstream tasks for a long time.
Several recent works have leveraged NLP methods and simple sampling methods for different downstream results 
The study of whether existing LMs can understand instructions has motivated a range of subsequent works. 
\citet{mishra-etal-2022-cross} proposed natural language instructions for cross-task generalization of LMs. 
PromptSource and FLAN \cite{wei2022finetuned} were built to leverage instructions and achieve zero-shot generalization on unseen tasks. 
Moreover, \citet{parmar-etal-2022-boxbart} shows the effectiveness of instructions in multi-task settings for the biomedical domain. 
\citet{mishra-etal-2022-reframing} discussed the impact of task instruction reframing on model response.
\citet{gupta2022john} showed that adding knowledge with instruction helps LMs understand the context better.
Furthermore, several approaches have been proposed to improve model performance using instructions, including \cite{wang-etal-2022-super, luo2022biotabqa, mishra2022help} 
Several studies are present that show adding knowledge with instruction helps LMs understand the context better \cite{gupta2021context}.

\section{Additional Tables for Plots}

The following section presents the absolute non aggregated numbers for the plots generated to analyse the instruction effectiveness (Figure \ref{fig:eff_instr}) as well as the sample efficiency plots (Figure \ref{fig:sample_eff}). 
The following analysis was conducted on the 3 subtasks viz. ATE, ATSC and AOSTE. 
This was based on the level of difficulty of the tasks. 
To balance out the analysis across tasks of various difficulties, we chose the easiest task which is just task extraction. 
It was followed by ATSC task which is more complicated since the model has to learn associations of the aspect term and its corresponding sentiment polarity. 
Finally the task with maximum difficulty was triplet extraction since the model has to extract all triplets given a sentence.

Table \ref{tab:instr_eff_ate} presents the performance metrics in terms of F1 score for the ATE subtask for the 4 datasets when instruction tuned with various configuration of instructions as mentioned in \S \ref{analysis}.
Similarly Table \ref{tab:instr_eff_atsc} presents the F1 scores for the ATSC subtask when instruction tuned with various configuration of instructions as mentioned in \S \ref{analysis}. 
Table \ref{tab:instr_eff_aoste} presents the F1 scores for the AOSTE subtask when instruction tuned with various configuration of instructions as mentioned in \S \ref{analysis}. 
Finally, Table \ref{tab:sample_efficiency}, describes the values for the sample efficiency plot. This plot presents the raw unnagregated numbers for ATE, ATSC and AOSTE.

\section{\name{} prompt examples}
\label{sec:prompt_examples}

 \begin{figure}[t!]
	\includegraphics[width=7.4 cm, height= 8.4 cm]{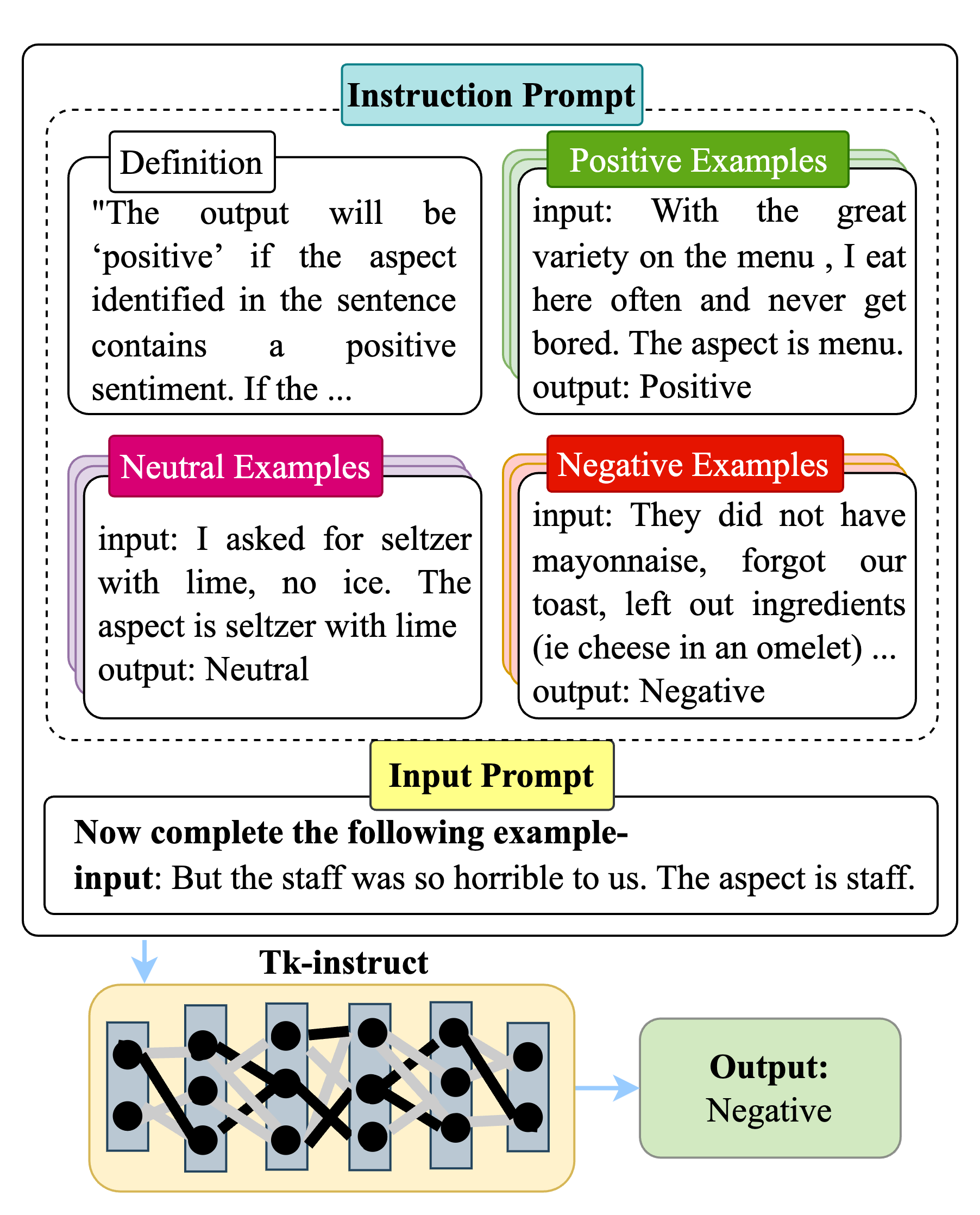}
	\caption{Formulation of \name{} for ATSC task. The input consists of an instruction prompt and a sentence. The output label is the sentiment polarity for the corresponding aspect.}
	\label{fig:flowchart}
\end{figure} 

The instruction prompts for \name{}-1, and \name{}-2 are presented in detail for all three ABSA subtasks. 
Table \ref{tab:ate_ip}, \ref{tab:atsc_ip}, and \ref{tab:jt_ip} presents the prompts provided for \name{}-2 model for the ATE, ATSC, and AOPE, respectively. 

For the \name{}-1 model, the instruction prompts are similar, with the difference that negative and neutral examples are not provided in the instruction prompts.

\input{tables/instr_eff_ate}

\input{tables/instr_eff_atsc}

\input{tables/instr_eff_aoste}

\input{tables/sample_efficiency_tables}


\input{tables/absa2_ate_example.tex}

\input{tables/absa2_atsc_example.tex}

\input{tables/absa2_joint_example.tex}

\input{tables/absa2_aooe_example}

\input{tables/absa2_aope_example}

\input{tables/absa2_aoste_example}

\input{tables/if_taskdef_aope}

\input{tables/if_absa3_aope}

\input{tables/if_absa4_aope}

\input{tables/if_absa5_aope}

\input{tables/if_absa6_aope}

\end{document}

%% file: tables/ATERes.tex
\begin{table}[t!]
\centering
\resizebox{\columnwidth}{!}
{
\begin{tabular}{lrrrr}
\hline
\textbf{Model} & \textbf{Lapt14} & \textbf{Rest14} & \textbf{Rest15} & \textbf{Rest16} \\ \hline
GPT2$_{med}$  & 82.04 & 75.94 & - & - \\
GRACE                & 87.93 & 85.45 & - & - \\ 
BARTABSA                & 83.52 & 87.07 & 75.48 & - \\ 
IT-MTL                & 76.93 & - & 74.03 & 79.41 \\ \hline
\name{}1 & 91.40 & \textbf{92.76} & 75.23 & \textbf{81.48} \\
\name{}2 & \textbf{92.30} & 92.10 & \textbf{76.64} & 80.32 \\ \hline
\end{tabular}
}
\caption{ATE subtask results denoting F1 scores. GPT2$_{med}$, GRACE, BARTABSA and IT-MTL results are from \citet{hosseini-asl-etal-2022-generative}, \citet{luo-etal-2020-grace}, \citet{yan-etal-2021-unified} and \citet{varia2023instruction} respectively.}

\label{tab:ate_result}
\end{table}

%% file: tables/ATSCRes.tex
\begin{table}[t!]
\centering
\resizebox{\columnwidth}{!}
{
\begin{tabular}{lrrrr}
\hline
\textbf{Model} & \textbf{Lapt14} & \textbf{Rest14} & \textbf{Rest15} & \textbf{Rest16} \\ \hline
ABSA-DeBERTa        & 82.76 & 89.46 & - & - \\
LSAT       & \textbf{86.31} & \textbf{90.86} & - & - \\ 
Dual-MRC          & 75.97 & 82.04 & 73.59 & - \\ \hline
\name{}1 & 80.62 & 86.25 & 83.02 & 89.10 \\
\name{}2 & 81.56 & 85.17 & \textbf{84.50} & \textbf{89.43} \\ \hline
\end{tabular}
}
\caption{ATSC subtask results denoting accuracy. ABSA-DeBERTa, LSAT and dual-MRC are from \citet{Marcacini2021AspectbasedSA}, \citet{yang2021improving} and \citet{mao2021joint} respectively.}
\label{tab:atsc_result}
\end{table}

%% file: tables/JointTaskRes.tex
\begin{table}[t!]
\centering
\resizebox{\columnwidth}{!}
{
\begin{tabular}{lrrrr}
\hline
\textbf{Model} & \textbf{Lapt14} & \textbf{Rest14} & \textbf{Rest15} & \textbf{Rest16} \\ \hline
GRACE                       & 75.97     & 78.07     & -     & - \\ 
BARTABSA                       & 67.37     & 73.56     & 66.61     & - \\ 
IT-MTL                       & 66.07     & -      & 67.06    & 74.07 \\ 
\hline
\name{}1 & 78.89 & 76.16 & 69.02 & \textbf{74.24} \\
\name{}2 & \textbf{79.34} & \textbf{79.47} & \textbf{69.39} & 73.06 \\
\hline
\end{tabular}
}
\caption{ASPE subtask results denoting F1 scores. 
GRACE, BARTABSA and IT-MTL results are from \citet{luo-etal-2020-grace}, \citet{yan-etal-2021-unified} and \citet{varia2023instruction}.}
\label{tab:joint_result}
\end{table}

%% file: tables/AOOE_res.tex
\begin{table}[t!]
\centering
\resizebox{\columnwidth}{!}
{
\begin{tabular}{lrrrr}
\hline
\textbf{Model} & \textbf{Lapt14} & \textbf{Rest14} & \textbf{Rest15} & \textbf{Rest16} \\ \hline
IOG & 70.99  & 80.23    & 71.91   & 81.60 \\
ONG   & 76.77     & 82.33    & 78.81   & 86.01 \\
BARTABSA & \textbf{80.55}  & \textbf{85.38}    & 80.52   & \textbf{87.92} \\
\hline
\name{}1 & 76.42 & 80.78 & 80.41 & 83.07 \\
\name{}2 & 77.16 &81.08 & \textbf{81.34} & 83.27 \\
\hline
\end{tabular}
}
\caption{Results of the AOOE subtask denoting F1 scores. 
IOG, ONG and BARTABSA are from \citet{fan-etal-2019-target}, \citet{pouran-ben-veyseh-etal-2020-introducing} and \citet{yan-etal-2021-unified} respectively.
}
\label{tab:aooe_result}
\end{table}

%% file: tables/AOPE_res.tex
\begin{table}[t!]
\centering
\resizebox{\columnwidth}{!}
{
\begin{tabular}{lrrrr}
\hline
\textbf{Model} & \textbf{Lapt14} & \textbf{Rest14} & \textbf{Rest15} & \textbf{Rest16} \\ \hline
Seq2Path   & \textbf{74.29}     & \textbf{77.35}     &\textbf{71.84}     & \textbf{79.09} \\
GAS              & 69.55     & 75.15     & 67.93     & 75.42 \\ 
BMRC   & 67.45     & 76.23     & 68.60     & 76.52 \\ 
\hline
\name{}1 & 60.75 & 70.46 & 60.31 & 72.04 \\
\name{}2 & 61.74 & 71.37 & 62.59 & 70.06 \\
\hline
\end{tabular}
}
\caption{Results of the AOPE subask denoting F1 scores. 
Seq2Path, GAS and BMRC are from \citet{mao-etal-2022-seq2path}, \citet{zhang-etal-2021-towards-generative} and \citet{chen2021bidirectional} respectively.
}
\label{tab:aope_result}
\end{table}

%% file: tables/AOSTE_res.tex
\begin{table}[t!]
\centering
\resizebox{\columnwidth}{!}
{
\begin{tabular}{lrrrr}
\hline
\textbf{Model} & \textbf{Lapt14} & \textbf{Rest14} & \textbf{Rest15} & \textbf{Rest16} \\ \hline
BMRC   & 59.27      & 70.69     & 61.05     & 68.13 \\
Seq2Path  & \textbf{65.27} & \textbf{75.52} & \textbf{65.88} & \textbf{73.67} \\
IT-MTL    & -     & 43.84   & 52.94    & 53.75 \\ 
\hline
\name{}1 & 60.67 & 70.50 & 60.63 & 68.15 \\
\name{}2 & 61.86 & 71.17 & 59.98 & 70.72 \\
\hline
\end{tabular}
}
\caption{Results of the AOSTE subask denoting F1 scores. 
Seq2Path, GAS and BMRC are from \citet{chen2021bidirectional}, \citet{mao-etal-2022-seq2path} and \citet{varia2023instruction}.
}
\label{tab:aoste_result}
\end{table}

%% file: tables/crossdomainRes.tex
\begin{table}[t!]
\centering
\resizebox{\linewidth}{!}
{
\begin{tabular}{lll|rrr}
\hline
\multicolumn{1}{c}{\textbf{Train}} & \multicolumn{1}{c}{\textbf{Test}} & \multicolumn{1}{c|}{\textbf{Model}} & \multicolumn{1}{c}{\textbf{ATE}} & \multicolumn{1}{c}{\textbf{ATSC}} & \multicolumn{1}{c}{\textbf{ASPE}} \\ \hline
\multirow{2}{*}{Rest14}            & \multirow{2}{*}{Lapt14}           & \name{}-1                                  & 71.98                            & 80.56                             & 64.30                              \\
                                   &                                   & \name{}-2                                  & \multicolumn{1}{l}{71.83}        & 82.44                             & 65.30                              \\ \hline
\multirow{2}{*}{Lapt14}            & \multirow{2}{*}{Rest14}           & \name{}-1                                  & 62.85                            & 75.53                             & 55.06                              \\
                                   &                                   & \name{}-2                                  & 76.85                            & 80.56                             & 62.95                              \\ \hline
\multirow{2}{*}{Rest15}            & \multirow{2}{*}{Hotel15}           & \name{}-1                                  & 74.51                            & 87.65                             & 66.88                              \\
                                   &                                   & \name{}-2                                  & 70.53                            & 89.74                             & 67.82                              \\ \hline
\end{tabular}
}
\caption{Results of the cross-domain evaluation where the model is trained on Lapt14 and the test set is of Rest14 and vice versa. The results of the model trained on Rest15 and evaluated on Hotel15 is also reported.}
\label{tab:cross_domain}
\end{table}

%% file: tables/cross_domain_joint.tex
\begin{table}[t!]

\resizebox{\linewidth}{!}
{\begin{tabular}{llrrr}
\hline
\multicolumn{1}{c}{\textbf{Task}} & \textbf{Model} & \multicolumn{1}{c}{\textbf{ATE}} & \multicolumn{1}{c}{\textbf{ATSC}} & \multicolumn{1}{c}{\textbf{ASPE}} \\ \hline
\multirow{2}{*}{Lapt14} & \name{}-1 & 90.35 & 81.09   & 80.07 \\
                        & \name{}-2 & 93.28 & 83.60  & 80.47  \\ \hline
\multirow{2}{*}{Rest14} & \name{}-1 & 88.88 & 86.42  & 80.81 \\
                        & \name{}-2 & 93.55 & 88.03  & 79.70  \\ \hline
\end{tabular}}
\caption{Results of joint-domain evaluation where the model is trained on both Lapt14 and Rest14 datasets and evaluated on the respective test set.}
\label{tab:cross_domain_joint}
\end{table}

%% file: tables/LoraComparison.tex
\begin{table}[t!]
\centering
\resizebox{\columnwidth}{!}
{
\begin{tabular}{l|ll|ll|ll}
\hline
\multicolumn{1}{c|}{\multirow{2}{*}{\textbf{Tasks}}} & \multicolumn{2}{c|}{\textbf{ATE}} & \multicolumn{2}{c|}{\textbf{ATSC}} & \multicolumn{2}{c}{\textbf{ASPE}} \\ \cline{2-7} 
\multicolumn{1}{c|}{}                  & Lapt14      & Rest14     & Lapt14      & Rest14      & Lapt14      & Rest14     \\ \hline
LoRA 8                              & 73.51       & 79.43      & 55.79       & 59.08       & 53.19       & 57.28      \\
LoRA 16                             & 73.57       & 78.32      & 54.30       & 59.16       & 52.30       & 57.19      \\
LoRA 32                             & 75.52       & 78.74      & 54.94       & 59.58       & 54.43       & 56.98      \\
LoRA 64                              & 71.61       & 76.93      & 55.87       & 58.64       & 55.87       & 58.64      \\
\name{}-1    & 91.40      & 92.76            & 80.62      & 86.25      & 78.89       & 76.16            \\
\name{}-2   & 92.30            & 92.10            &81.56     & 85.17              & 79.34            & 79.47            \\ \hline
\end{tabular}
}
\caption{Results of LoRA PEFT and \name{}-1 and \name{}-2 across all subtasks. 8, 16, 32 and 64 in LoRA denote the rank of the adapter method.}
\label{tab:lora_comparison}
\end{table}

%% file: tables/ATSCData.tex
\begin{table}[H]
\centering
\begin{tabular}{clrrr}
\hline
Dataset               & \multicolumn{1}{c}{Split} & \multicolumn{1}{c}{Pos.} & \multicolumn{1}{c}{Neg.} & \multicolumn{1}{c}{Neut.} \\ \hline
Lapt14               & Train          & 987        & 866               & 460                         \\
                      & Test          & 341         & 128               & 169                         \\ 
                      \hline

Rest14           & Train            & 2164         & 805              & 633                         \\
                 & Test         & 728          & 196              & 196                         \\ 
                    \hline

Rest15           & Train            & 912         & 256              & 36                         \\
                 & Test         & 326          & 182              & 34                         \\ 
                    \hline

Hotel15          & Test         & 163          & 45              & 7                         \\ 
                    \hline

Rest16           & Train            & 1240         & 439              & 69                         \\
                 & Test         & 468          & 117              & 30                         \\ 
                    \hline
\end{tabular}

\caption{Dataset Statistics for ATSC subtask denoting number of samples. Pos., Neg., and Neut. represent Positive, Negative, and Neutral, respectively}
\label{tab:dataset_description_2}
\end{table}

%% file: tables/JTData.tex
\begin{table*}[]
\centering
\resizebox{\textwidth}{!}
{
\begin{tabular}{l|l|c|c|c|c|c|c|c|c|c|c|c|c}
\hline
Dataset     & Split & \multicolumn{1}{l|}{\#NO} & \multicolumn{1}{l|}{\#1} & \multicolumn{1}{l|}{\#2} & \multicolumn{1}{l|}{\#3} & \multicolumn{1}{l|}{\#4} & \multicolumn{1}{l|}{\#5} & \multicolumn{1}{l|}{\#6} & \multicolumn{1}{l|}{\#7} & \multicolumn{1}{l|}{\#8} & \multicolumn{1}{l|}{\#9} & \multicolumn{1}{l|}{\#10+} & \multicolumn{1}{l}{\#Total} \\ \hline
Lapt14     & Train & 1557                       & 930                       & 354                       & 140                       & 43                        & 10                        & 6                         & 3                         & 1    & -         & 1     & 3045                        \\
            & Test  & 378                        & 266                       & 105                       & 34                        & 10                        & 6                         & 1                         & -    & -                         & -                         & -                          & 800                         \\ \hline
            
Rest14 & Train & 1020                       & 1022                      & 572                       & 269                       & 104                       & 30                        & 15                        & 5                         & 3             & 1                     & -                          & 3041                        \\
            & Test  & 194                        & 290                       & 186                       & 80                        & 30                        & 14                        & 3                 & 2                         & -                         & -                         & 1    & 800                         \\ \hline

Rest15 & Train & 482                       & 576                      & 174                       & 58                       & 22                       & 2                        & -                        & -                         & 1    & -                    & -                          & 1315                        \\
    & Test  & 284                        & 294                       & 82                       & 18                        & 6                        & -                        & 1                         & -                         & -                         & -                         & -     & 685                         \\ \hline

Hotels15 & Test  & 98                        & 135                       & 23                       & 7                        & 2                        & 1                        & -                         & -                         & -                         & -                         & -     & 266                         \\ \hline

Rest16 & Train & 766                & 868                      & 258                       & 76                       & 28                       & 2                        & 1                        & -                      & 1    & -    & -                         & 2000                        \\
    & Test  & 256                        & 298                       & 87                       & 22                        & 9                        & 3                        & -                         & -                         & -                         & -                         & 1     & 676                         \\ \hline

\end{tabular}
}
\caption{Count of Aspects for the ATE, ASTE, AOOE, AOPE and AOSTE subtasks. \#$k$ is the count of samples that have $k$ aspects/aspect-sentiment polarity pairs in them. \#NO is the number of samples that have no aspect/aspect-sentiment polarity pairs in them.}
\label{tab:dataset_description_1}
\end{table*}

%% file: tables/instr_eff_ate.tex
\begin{table}[H]
\centering
\resizebox{\columnwidth}{!}
{
\begin{tabular}{lrrrr}
\hline
\multicolumn{1}{c}{\begin{tabular}[c]{@{}c@{}}Instruction\\ Type\end{tabular}} & \multicolumn{1}{l}{Lapt14} & \multicolumn{1}{l}{Rest14} & \multicolumn{1}{l}{Rest15} & \multicolumn{1}{l}{Rest16} \\ \hline
vanilla\_t5                                                                    & 71.67                      & 74.59                      & 61.74                      & 74.04                      \\
instruct\_t5                                                                   & 73.02                      & 77.25                      & 63.90             & 75.04                      \\
vanilla\_tk                                                                    & 83.07                      & 85.23             & 70.40                      & 78.04                      \\
task\_def\_only                                                                & 85.60                       & 86.78             & 72.31                      & 78.32                      \\
absa1                                                                          & 91.40                       & 92.76                      & 75.23                      & 81.48                      \\
absa2                                                                          & 92.30                       & 92.10              & 76.64                      & 80.32                      \\
absa3                                                                          & 88.06                      & 89.19                      & 72.31                      & 74.52                      \\
absa4                                                                          & 87.25                      & 87.78             & 71.81                      & 71.81                      \\
absa5                                                                          & 85.58                      & 86.00                      & 70.35                      & 68.33                      \\
absa6                                                                          & 83.91                      & 84.21                      & 68.89             & 64.85     \\ \hline                  
\end{tabular}
}
\caption{Tabular Results Instruction Effectiveness Plot for ATE}
\label{tab:instr_eff_ate}
\end{table}

%% file: tables/instr_eff_atsc.tex
\begin{table}[H]
\centering
\resizebox{\columnwidth}{!}
{
\begin{tabular}{lrrrr}
\hline
\multicolumn{1}{c}{\begin{tabular}[c]{@{}c@{}}Instruction\\ Type\end{tabular}} & \multicolumn{1}{l}{Lapt14} & \multicolumn{1}{l}{Rest14} & \multicolumn{1}{l}{Rest15} & \multicolumn{1}{l}{Rest16} \\ \hline
vanilla\_t5                                                                    & 59.42                      & 80.70                      & 72.41                      & 81.44                      \\
instruct\_t5                                                                   & 62.56                      & 81.30                      & 74.03                      & 82.54                      \\
vanilla\_tk                                                                    & 71.98                      & 83.10                      & 78.91                      & 85.86                      \\
task\_def\_only                                                                & 74.56                      & 83.27                      & 80.12                      & 86.45                      \\
absa1                                                                          & 79.37                      & 85.15                      & 82.98                      & 89.09                      \\
absa2                                                                          & 80.84                      & 84.47                      & 83.37                      & 88.66                      \\
absa3                                                                          & 79.01                      & 82.34                      & 81.67                      & 87.12                      \\
absa4                                                                          & 77.18                      & 80.21                      & 79.97                      & 85.58                      \\
absa5                                                                          & 75.35                      & 78.08                      & 78.27                      & 84.04                      \\
absa6                                                                          & 70.12                      & 75.95                      & 76.57                      & 82.50            \\\hline            
               
\end{tabular}
}
\caption{Tabular Results Instruction Effectiveness Plot for ATSC}
\label{tab:instr_eff_atsc}
\end{table}

%% file: tables/instr_eff_aoste.tex
\begin{table}[H]
\centering
\resizebox{\columnwidth}{!}
{
\begin{tabular}{lrrrr}
\hline
\multicolumn{1}{c}{\begin{tabular}[c]{@{}c@{}}Instruction\\ Type\end{tabular}} & \multicolumn{1}{l}{Lapt14} & \multicolumn{1}{l}{Rest14} & \multicolumn{1}{l}{Rest15} & \multicolumn{1}{l}{Rest16} \\ \hline
vanilla\_t5                                                                    & 53.53                      & 66.48                      & 64.53                      & 52.73                      \\
instruct\_t5                                                                   & 54.72                      & 67.15                      & 63.88                      & 55.30                       \\
vanilla\_tk                                                                    & 58.29                      & 69.16                      & 61.93                      & 63.01                      \\
task\_def\_only                                                                & 59.48                      & 69.83                      & 61.28                      & 65.58                      \\
absa1                                                                          & 60.67                      & 70.50                       & 60.63                      & 68.15                      \\
absa2                                                                          & 61.86                      & 71.17                      & 59.98                      & 70.72                      \\
absa3                                                                          & 58.98                      & 69.65                      & 57.83                      & 69.12                      \\
absa4                                                                          & 56.10                       & 68.13                      & 55.68                      & 67.52                      \\
absa5                                                                          & 53.22                      & 66.61                      & 53.53                      & 65.92                      \\
absa6                                                                          & 50.34                      & 65.09                      & 51.38                      & 64.32  \\\hline               
\end{tabular}
}
\caption{Tabular Results Instruction Effectiveness Plot for AOSTE}
\label{tab:instr_eff_aoste}
\end{table}

%% file: tables/sample_efficiency_tables.tex
\begin{table}[H]
\centering
\resizebox{\columnwidth}{!}
{
\begin{tabular}{lrrr}
\hline
Task  & \multicolumn{1}{c}{\begin{tabular}[c]{@{}c@{}}Sample \\ Size\end{tabular}} & \multicolumn{1}{l}{No Instruction} & \multicolumn{1}{l}{InstructABSA-2} \\ \hline
ate   & 10                                                                         & 49.15                              & 71.81                              \\
ate   & 20                                                                         & 56.12                              & 74.06                     \\
ate   & 50                                                                         & 68.30                      & 82.37                              \\
ate   & 100                                                                        & 73.13                     & 92.20                               \\
atsc  & 10                                                                         & 37.24                              & 49.67                              \\
atsc  & 20                                                                         & 51.23                     & 62.34                              \\
atsc  & 50                                                                         & 63.45                              & 73.21                              \\
atsc  & 100                                                                        & 70.06                     & 82.65                             \\
aoste & 10                                                                         & 26.34                              & 48.98                              \\
aoste & 20                                                                         & 45.78                              & 59.24                     \\
aoste & 50                                                                         & 54.29                              & 63.25                              \\
aoste & 100                                                                        & 60.05                     & 67.16                              \\ \hline
\end{tabular}
}
\caption{Tabular Results of Sample Efficiency Plots}
\label{tab:sample_efficiency}
\end{table}

%% file: tables/absa2_ate_example.tex
\begin{table*}[]
\begin{tabular}{ll}
\hline
\textbf{Task}             & Aspect Term Extraction (ATE)                                                \\ \hline
\textbf{Definition}       & Definition: The output will be the aspects (both implicit and explicit)           \\
\textbf{}                 & which have an associated opinion that is extracted from the input text.          \\
\textbf{}                 & In cases where there are no aspects, the output should be noaspectterm.            \\ \hline
\textbf{Positive }        &  Example Input 1: With the great variety on the menu, I eat here often and never get bored. \\
\textbf{Example}          &  Example Output 1: menu                                                                      \\
\textbf{}                 &  Example Input 2: Great food, good size menu, great service, and an unpretentious setting.    \\
\textbf{}                 &  Example output 2: food, menu, service, setting                                              \\ \hline
\textbf{Negative }        & Negative input 1: They did not have mayonnaise, forgot our toast,                            \\
\textbf{Example}                 & left out ingredients...\\
\textbf{}                 & Negative output 1: toast, mayonnaise, bacon, ingredients, plate                              \\
\textbf{}                 & Negative input 2: The seats are uncomfortable if you are sitting against the wall            \\
\textbf{}                 & on wooden benches.                                                                \\
\textbf{}                 & Negative output 2: seats                                                                     \\ \hline
\textbf{Neutral }                 & Neutral Input 1: I asked for a seltzer with lime, no ice.                                     \\
\textbf{Example}                 & Neutral Output 1: seltzer with lime                                                         \\
\textbf{}                 & Neutral Input 2: They wouldn't even let me finish my glass of wine before offering another.  \\
\textbf{}                 & Neutral Output 2: glass of wine                                                             \\ \hline
\textbf{Input}            & Now complete the following example-                                               \\
\textbf{}                 & input: My son and his girlfriend both wanted cheeseburgers and they were huge!    \\
                          & output: cheeseburgers                                                              \\ \hline
\end{tabular}
\caption{Illustrating \name{}-2 instruction prompting for the ATE sub task.}
\label{tab:ate_ip}
\end{table*}

%% file: tables/absa2_atsc_example.tex
\begin{table*}[]
\begin{tabular}{ll}
\hline
\textbf{Task}             & Aspect Term Sentiment Classification (ATSC)                                                  \\ \hline
\textbf{Definition} & The output will be 'positive', 'negative' or 'neutral' if the sentiment of the \\
\textbf{}           & identified aspect in the input is positive, negative or neutral respectively \\
\textbf{}           & For the aspects which are classified as noaspectterm, the sentiment is none.                  \\ \hline
\textbf{Positive}           & Example Input 1: With the great variety on the menu, I eat here often and never get bored.\\
\textbf{Example}           & Aspect: menu\\
\textbf{}                 & Example Output 1: positive                                                        \\
\textbf{}           & Example Input 2: Great food, good size menu, great service, and an unpretentious setting.\\
\textbf{}           & Aspect: food.\\
\textbf{}                 & Example Output 2: positive                                                        \\ \hline
\textbf{Negative}           & Example Input 1: They did not have mayonnaise, forgot our toast,  left out ingredients                   \\
\textbf{Example}                 & (i.e., cheese in an omelet), below hot temperatures and the bacon was      \\
\textbf{}           & so overcooked it crumbled on the plate when you touched it. Aspect: toast                     \\
\textbf{}                 & Example Output 1: negative                                                        \\
\textbf{}                 & Example Input 2: The seats are uncomfortable if you are sitting against the wall  \\
\textbf{}                 & on wooden benches. Aspect: seats                                        \\
\textbf{}                 & Example Output 2: negative                                                        \\ \hline
\textbf{Neutral}                 & Example Input 1: I asked for a seltzer with lime, no ice. Aspect: seltzer with lime \\
\textbf{Example}                 & Example Output 1: neutral                                                         \\
\textbf{}           & Example Input 2: They wouldn't even let me finish my glass of wine before offering another.               \\
                          & Aspect: a glass of wine                                                   \\
\textbf{}                 & Example Output 2: neutral                                                         \\ \hline
\textbf{Input}            & Now complete the following example-                                     \\
\textbf{}           & input: My son and his girlfriend both wanted cheeseburgers and they were huge!                 \\
                          & Aspect: cheeseburgers.                                                  \\
                          & output: positive                                                                \\ \hline
\end{tabular}
\caption{Illustrating \name{}-2 instruction prompting for the ATSC subtask.}
\label{tab:atsc_ip}
\end{table*}

%% file: tables/absa2_joint_example.tex
\begin{table*}[]
\begin{tabular}{ll}
\hline
\textbf{Task}             & Aspect Sentiment Pair Extraction (ASPE)                                             \\ 
\hline
\textbf{Definition} & Definition: The output will be the aspects (both implicit and explicit), and the aspects \\
                          & sentiment polarity. In cases where there are no aspects, the output \\
\textbf{}                 & should be no aspect-tern: none.                                       \\ \hline
\textbf{Positive}           & Example Input 1: With the great variety on the menu, I eat here often and never get bored.       \\
\textbf{Example}                 & Example Output 1: menu:positive                                              \\
\textbf{}           & Example Input 2: Great food, good size menu, great service, and an unpretentious setting.          \\
\textbf{}                 & Example Output 2: food:positive                                              \\ \hline
\textbf{Negative}           & Example Input 1: They did not have mayonnaise, forgot our toast,  left out ingredients            \\
\textbf{Example}           & (i.e., cheese in an omelet), below hot temperatures, and the bacon was                      \\
\textbf{}                 & so overcooked it crumbled on the plate when you touched it.       \\
\textbf{}                 & Example Output 1: toast:negative                                             \\
\textbf{}           & Example Input 2: The seats are uncomfortable if you are sitting against the wall                  \\
\textbf{}                 & on wooden benches. Aspect: seats                                   \\
\textbf{}                 & Example Output 2: negative                                                   \\ \hline
\textbf{Neutral}                 & Example Input 1: I asked for a seltzer with lime, no ice.                      \\
\textbf{Example}                 & Example Output 1: seltzer with lime: neutral                                  \\
\textbf{}                 & Example Input 2: They wouldn't even let me finish my glass of wine before     \\
                          & offering another.                                                  \\
\textbf{}                 & Example Output 2: glass of wine:neutral                                      \\ \hline
\textbf{Input}            & Now complete the following example-                                \\
\textbf{}                 & input: My son and his girlfriend both wanted cheeseburgers and they were huge!      \\
                          & output: cheeseburgers: positive                                                            \\ \hline
\end{tabular}
\caption{
     Illustrating \name{}-2 instruction prompting for the ASPE subtask.}
\label{tab:jt_ip}
\end{table*}

%% file: tables/absa2_aooe_example.tex
\begin{table*}[]
\resizebox{\linewidth}{!}
{
\begin{tabular}{ll}
\hline
\textbf{Task} & Aspect Oriented Opinion Extraction (AOOE) \\ 
\hline                                                  
\textbf{Definition} & Definition: The output will be the opinion/describing word of the aspect terms in the \\
& sentence. In cases where there are no aspects the output should be none. \\ 
\hline                                          
\textbf{Positive}   & Example Input 1: Faan 's got a great concept but a little rough on the delivery. \\ 
\textbf{Example}    &  Example Output 1: delivery:rough           \\
\textbf{}           & Example Input 2: it is of high quality , has a killer GUI , is extremely stable,\\
\textbf{}           &is highly expandable. The aspect is GUI.\\
\textbf{}           & Example Output 2: killer                                              \\ 
\hline                             
\textbf{Negative}   & Example Input 1: One night I turned the freaking thing off after using it , the next day \\
\textbf{Example}    & I turn it on , no GUI , screen all dark,.. The aspect is GUI.                      \\
\textbf{}           & Example Output 1: no                                             \\
\textbf{}           & Example Input 2: I can barely use any usb devices because they will     \\
\textbf{}           & not stay connected properly . The aspect is usb devices.                \\
\textbf{}           & Example Output 2: not stay connected properly                     \\ 
\hline
\textbf{Neutral}    & Example Input 1: However, ..external mouse unnecessary. The aspect is external mouse. \\
\textbf{Example}    & Example Output 1: unnecessary         \\                                        
\textbf{}           & Example Input 2: ... extended warranty and they refused. The aspect is extended warranty. \\ 
\textbf{}           & Example Output 2: refused                                   \\ 
\hline
\textbf{Input}      & Now complete the following example-                                \\
\textbf{}           & input: My son ... cheeseburgers and they were huge!. The aspect is cheeseburgers. \\
                    & output: huge\\ \hline
\end{tabular}
}
\caption{
     Illustrating \name{}-2 instruction prompting for the AOOE subtask.}
\label{tab:aooe_ip}
\end{table*}

%% file: tables/absa2_aope_example.tex
\begin{table*}[]
\resizebox{\linewidth}{!}
{
\begin{tabular}{ll}
\hline
\textbf{Task} & Aspect Opinion Pair Extraction (AOPE) \\ 
\hline                                                  
\textbf{Definition} & Definition: The output will be the aspect terms in the \\
& sentence followed by its describing/opinion term.\\ 
\hline                                          
\textbf{Positive}   & Example Input 1: I charge it at night and skip taking the cord with me because of the \\ 
\textbf{Example}    &  good battery life.       \\
\textbf{}           & Example Output 1: battery life:good   \\
\textbf{}           & Example Input 2: it is of high quality , has a killer GUI , is extremely stable,\\
\textbf{}           &is highly expandable,.. good applications,.. easy to use.\\
\textbf{}           & Example Output 2: quality:high, GUI:killer, applications:good, use:easy  \\ 
\hline                             
\textbf{Negative}   & Example Input 1: A month or so ago , the freaking motherboard just died . \\
\textbf{Example}    & Example Output 1: motherboard:freaking \\
\textbf{}           & Example Input 2: I had always used PCs ....crashing and the poorly designed \\
\textbf{}           & operating systems that were never very intuitive               \\
\textbf{}           & Example Output 2: operating systems:poorly designed, operating systems: never very intuitive  \\ 
\hline  
\textbf{Neutral}    & Example Input 1: It has a 10 hour ... when you 're doing web browsing and word editing , \\
\textbf{Example}    & making it perfect for the classroom or office, ...          \\
\textbf{}           & Example Output 1: web browsing:perfect, word editing:perfect          \\                            
\textbf{}           & Example Input 2: no complaints with their desktop , and maybe because it just sits \\ 
\textbf{}           &  on your desktop... which could jar the hard drive , or the motherboard     \\ 
\textbf{}           & Example Output 2: hard drive:jar, motherboard:jar                    \\ 
\hline 
\textbf{Input}      & Now complete the following example-                                \\
\textbf{}           & input: Boot time is super fast , around anywhere from 35 seconds to 1 minute \\
                    & output: Boot time:superfast\\ \hline
\end{tabular}
}
\caption{
     Illustrating \name{}-2 instruction prompting for the AOPE subtask.}
\label{tab:aope_ip}
\end{table*}

%% file: tables/absa2_aoste_example.tex
\begin{table*}[]
\resizebox{\linewidth}{!}
{
\begin{tabular}{ll}
\hline
\textbf{Task} & Aspect Opinion Sentiment Triplet Extraction (AOSTE) \\ 
\hline                                                  
\textbf{Definition} & Definition: The output will be the aspect terms in the \\
& sentence followed by their describing words and sentiment polarity.\\ 
\hline                                          
\textbf{Positive}   & Example Input 1: I charge it at night and skip taking the cord with me because of the \\ 
\textbf{Example}    &  good battery life.       \\
\textbf{}           & Example Output 1: battery life:good:positive   \\
\textbf{}           & Example Input 2: it is of high quality , has a killer GUI , is extremely stable,\\
\textbf{}           &is highly expandable,.. good applications,.. easy to use.\\
\textbf{}           & Example Output 2: quality:high:positive, GUI:kille:positive \\ 
\hline                             
\textbf{Negative}   & Example Input 1: A month or so ago , the freaking motherboard just died . \\
\textbf{Example}    & Example Output 1: motherboard:freaking \\
\textbf{}           & Example Input 2: I had always used PCs ....crashing and the poorly designed \\
\textbf{}           & OS that were never very intuitive               \\
\textbf{}           & Example Output 2: OS:poorly designed:negative, OS: never very intuitive:negative  \\ 
\hline  
\textbf{Neutral}    & Example Input 1: It has a 10 hour ... when you 're doing web browsing and word editing ,\\
\textbf{Example}    & making it perfect for the classroom or office, ...          \\
\textbf{}           & Example Output 1: web browsing:perfect:neutral, word editing:perfect:neutral \\        
\textbf{}           & Example Input 2: no complaints with their desktop , and maybe because it just sits \\ 
\textbf{}           &  on your desktop... which could jar the hard drive , or the motherboard     \\ 
\textbf{}           & Example Output 2: hard drive:jar:neutral, motherboard:jar:neutral                    \\ 
\hline 
\textbf{Input}      & Now complete the following example-                                \\
\textbf{}           & input: Boot time is super fast , around anywhere from 35 seconds to 1 minute \\
                    & output: Boot time:superfast:positive\\ \hline
\end{tabular}
}
\caption{
     Illustrating \name{}-2 instruction prompting for the AOPE subtask.}
\label{tab:aoste_ip}
\end{table*}

%% file: tables/if_taskdef_aope.tex
\begin{table*}[]
\resizebox{\linewidth}{!}
{
\begin{tabular}{ll}
\hline
\textbf{Task} & Aspect Opinion Pair Extraction (AOPE) - Task Definition Only \\ 
\hline                                                  
\textbf{Definition} & Definition: The output will be the aspect terms in the \\
& sentence followed by its describing/opinion term.\\ 
\hline                                          
\textbf{Input}      & Now complete the following example-                                \\
\textbf{}           & input: Boot time is super fast , around anywhere from 35 seconds to 1 minute \\
                    & output: Boot time:superfast\\ \hline
\end{tabular}
}
\caption{
     Illustrating Only Task Definition based prompting for AOPE subtask.}
\label{tab:task_def_aope}
\end{table*}

%% file: tables/if_absa3_aope.tex
\begin{table*}[]
\resizebox{\linewidth}{!}
{
\begin{tabular}{ll}
\hline
\textbf{Task} & Aspect Opinion Pair Extraction (AOPE) - 2 Negative Examples \\ 
\hline                                                  
\textbf{Definition} & Definition: The output will be the the aspect terms in the \\
& sentence followed by their describing/opinion term.\\ 
\hline                                                                    
\textbf{Negative}   & Example Input 1: A month or so ago , the freaking motherboard just died . \\
\textbf{Example}    & Example Output 1: motherboard:freaking:negative \\
\textbf{}           & Example Input 2: I had always used PCs ....crashing and the poorly designed \\
\textbf{}           & OS that were never very intuitive               \\
\textbf{}           & Example Output 2: OS:poorly designed, OS: never very intuitive  \\ 
\hline  
\end{tabular}
}
\caption{
     Illustrating Definition + 2 negative exemplars based prompting for AOPE subtask}
\label{tab:absa3_aope}
\end{table*}

%% file: tables/if_absa4_aope.tex
\begin{table*}[]
\resizebox{\linewidth}{!}
{
\begin{tabular}{ll}
\hline
\textbf{Task} & Aspect Opinion Pair Extraction (AOPE) - 2 Neutral Examples \\ 
\hline                                                  
\textbf{Definition} & Definition: The output will be the the aspect terms in the \\
& sentence followed by their describing/opinion term.\\ 
\hline                                                                    
\textbf{Neutral}    & Example Input 1: It has a 10 hour ... when you 're doing web browsing and word editing,\\
\textbf{Example}    & making it perfect for the classroom or office, ...          \\
\textbf{}           & Example Output 1: web browsing:perfect, word editing:perfect          \\                  
\textbf{}           & Example Input 2: no complaints with their desktop , and maybe because it just sits \\ 
\textbf{}           & on your desktop... which could jar the hard drive , or the motherboard     \\ 
\textbf{}           & Example Output 2: hard drive:jar, motherboard:jar                    \\ 
\hline 
\textbf{Input}      & Now complete the following example-                                \\
\textbf{}           & input: Boot time is super fast , around anywhere from 35 seconds to 1 minute \\
                    & output: Boot time:superfast\\ \hline
\end{tabular}
}
\caption{
     Illustrating Definition + 2 neutral exemplars based prompting for AOPE subtask}
\label{tab:absa4_aope}
\end{table*}

%% file: tables/if_absa5_aope.tex
\begin{table*}[]
\resizebox{\linewidth}{!}
{
\begin{tabular}{ll}
\hline
\textbf{Task} & Aspect Opinion Pair Extraction (AOPE) - 1 Positive, Negative and Neutral Example \\ 
\hline                                                  
\textbf{Definition} & Definition: The output will be the aspect terms in the \\
& sentence followed by its describing/opinion term.\\ 
\hline                                          
\textbf{Positive}   & Example Input 1: I charge it at night and skip taking the cord with me because of the \\ 
\textbf{Example}    &  good battery life.       \\
\textbf{}           & Example Output 1: battery life:good   \\
\hline                             
\textbf{Negative}   & Example Input 1: A month or so ago , the freaking motherboard just died . \\
\textbf{Example}    & Example Output 1: motherboard:freaking \\
\hline  
\textbf{Neutral}    & Example Input 1: It has a 10 hour ... when you 're doing web browsing and word editing , \\
\textbf{Example}    & making it perfect for the classroom or office, ...          \\
\textbf{}           & Example Output 1: web browsing:perfect, word editing:perfect          \\                 
\hline 
\textbf{Input}      & Now complete the following example-                                \\
\textbf{}           & input: Boot time is super fast , around anywhere from 35 seconds to 1 minute \\
                    & output: Boot time:superfast\\ \hline
\end{tabular}
}
\caption{
     Illustrating Definition + 1 positive + 1 negative + 1 neutral exemplars based prompting for AOPE subtask}
\label{tab:absa5_aope}
\end{table*}

%% file: tables/if_absa6_aope.tex
\begin{table*}[]
\resizebox{\linewidth}{!}
{
\begin{tabular}{ll}
\hline
\textbf{Task} & Aspect Opinion Pair Extraction (AOPE) - Delusive Examples \\ 
\hline                                                  
\textbf{Definition} & Definition: The output will be the aspect terms in the \\
& sentence followed by its describing/opinion term.\\ 
\hline                                          
\textbf{Positive}   & Example Input 1: I charge it at night and skip taking the cord with me because of the \\ 
\textbf{Example}    &  good battery life.       \\
\textbf{}           & Example Output 1: motherboard:freaking   \\
\hline                             
\textbf{Negative}   & Example Input 1: A month or so ago , the freaking motherboard just died . \\
\textbf{Example}    & Example Output 1: web browsing:perfect, word editing:perfect \\
\hline  
\textbf{Neutral}    & Example Input 1: It has a 10 hour ... when you 're doing web browsing and word editing , \\
\textbf{Example}    & making it perfect for the classroom or office, ...          \\
\textbf{}           & Example Output 1: battery life:good          \\                 
\hline 
\textbf{Input}      & Now complete the following example-                                \\
\textbf{}           & input: Boot time is super fast , around anywhere from 35 seconds to 1 minute \\
                    & output: Mac M1: fast\\ \hline
\end{tabular}
}
\caption{
     Illustrating delusive instruction based prompting for AOPE subtask. In this task, the output labels of the examplars are mapped incorrectly with the inputs.}
\label{tab:absa6_aope}
\end{table*}